\documentclass{article}
\usepackage{arxiv}

\usepackage{cite}
\usepackage{amsmath,amssymb,amsfonts}
\usepackage{algorithmic}
\usepackage{graphicx}
\usepackage{textcomp}
\usepackage{xcolor}
\usepackage{bm}
\usepackage{fontawesome}
\usepackage{authblk}
\newcommand{\etal}{\textit{et al.}}

\title{GazeVLM: A Vision-Language Model for Multi-Task Gaze Understanding}

\author{Athul M. Mathew\thanks{Corresponding author: \texttt{amathew@elm.sa}}}
\author{Haithem Hermassi}
\author{Thariq Khalid}
\author{Arshad Ali Khan}
\affil{Applied Research, Elm Company, Saudi Arabia}

\begin{document}

\maketitle

\begin{abstract}
Gaze understanding unifies the detection of people, their gaze targets, and objects of interest into a single framework, offering critical insights into visual attention and intent estimation. Prior methods have modelled gaze cues in visual scenes, but a unified system is still needed for gaze understanding using both visual and language prompts. This paper introduces GazeVLM, a novel Vision-Language Model (VLM) for multi-task gaze understanding in images, addressing person detection, gaze target detection, and gaze object identification. Unlike other transformer-based methods, GazeVLM represents the first application of a VLM to these combined tasks, allowing for selective execution of each task. Our ablation study on visual input combinations reveals that a fusion of RGB images with HHA-encoded depth maps, guided by text prompts, yields superior performance. We also introduce an object-level gaze detection metric for gaze object identification ($AP_{ob}$). Through experiments, GazeVLM demonstrates significant improvements, achieving state-of-the-art evaluation scores on GazeFollow and VideoAttentionTarget datasets.
\end{abstract}

\section{Introduction}
\label{sec:introduction}

Humans possess a remarkable ability to track eye gaze, enabling them to interpret focus and predict actions \cite{nips15_recasens}. Gaze offers critical insights into attention, cognition, and intent, serving as a powerful nonverbal cue that reveals visual focus and communicates interests \cite{tonini2023}. This has made gaze behaviour a key area of study across disciplines to better understand human attention and behaviour. Understanding human gaze is critical for developing intelligent systems capable of seamless human interaction. Teaching machines to accurately estimate gaze is a significant challenge \cite{tonini2023}.  

\begin{figure}[!ht]
  \centering
   \includegraphics[width=0.8\linewidth]{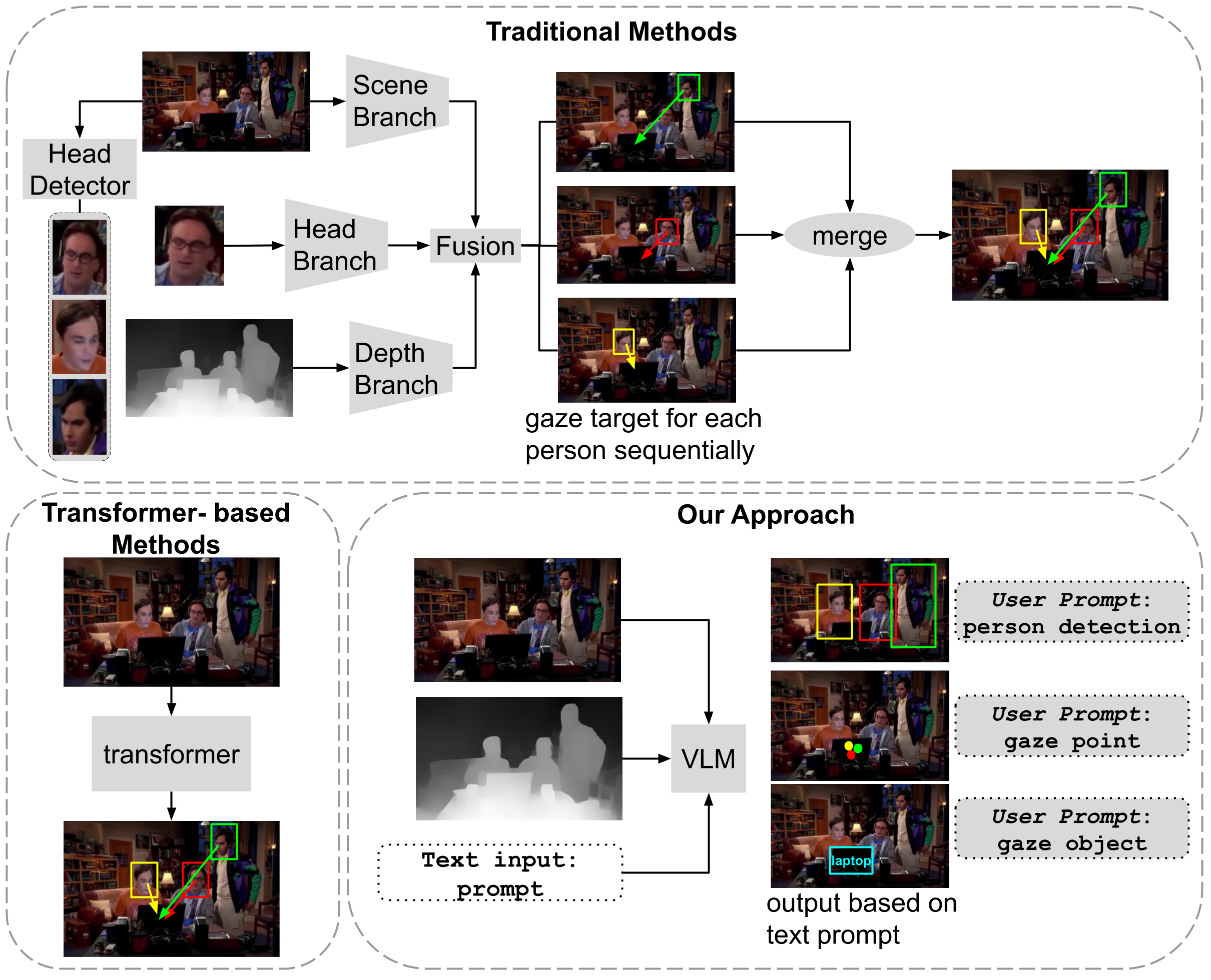}
   \caption{A comparison of previous methods versus our approach.}
   \label{fig:gaze_vlm_intro}
\end{figure}

Gaze understanding, the task of determining where and what a person looks at in everyday images, sits at the center of human-centric computer vision and directly impacts human-computer interaction, assistive perception, social signal analysis, and embodied AI~\cite{Tafasca_2024_CVPR,Gupta_2024_NeurIPS_MTGS}. In third-person imagery, progress hinges on solving three questions together: who is the subject, where in the scene the gaze lands, and what object or region receives attention. These questions are entangled because gaze depends on subtle ocular and facial cues while also being shaped by scene geometry and semantics. Treating them jointly is natural; doing so robustly is difficult.
Reliability in the wild remains difficult. Occlusions from hair or glasses, strong head-pose variation, and non-uniform lighting degrade facial evidence, while out-of-frame gaze targets break image-local heuristics \cite{ghosh2023automatic}. Error cascades are common: a near-miss in person localization shifts the crop, the head pose becomes slightly biased, and the downstream gaze heatmap drifts toward a salient but irrelevant object. The result looks plausible, but is wrong. These brittleness patterns motivate models that reason across appearance and geometry while preserving the interdependence among detection, target localization, and object identification. In this paper, we aim to bridge these three fundamental and intertwined tasks for gaze understanding using a single unified model - GazeVLM.

Recent works, such as Gaze-LLE \cite{ryan2024}, have explored the use of foundation models for gaze target detection. However, such methods do not incorporate person detection or gaze objects. Similar to prior methods, this research\cite{ryan2024} employs scene parsing, but with the key distinction that it makes use of DinoV2 transformer instead of CNN-based backbones used by the previous methods. Furthermore, Gaze-LLE\cite{ryan2024} requires an additional input—the head location—predicted by a separate model, which adds dependency and complexity to the pipeline.

Recently, Vision Language Models (VLM) are playing a big role in the convergence of Computer Vision and Large Language Models. They serve as a way to query and understand what is happening in images and videos using natural language as the foundation. VLMs are advanced AI models that blend computer vision and natural language processing, learning joint representations of images and text. VLMs excel at learning rich representations of visual and linguistic information, enabling them to perform a wide range of tasks, including image captioning, visual question answering, and object detection. There have been many generic VLMs as well as domain-specific VLMs in recent research spaces. 

VLMs have emerged as powerful tools for multimodal understanding, offering a promising new direction to address tasks. Recent efforts have explored integrating human gaze as an additional modality to align model attention with user intention \cite{yan2024voila}. For instance, a recent study demonstrated that prompting a VLM to describe scenes automatically captures relevant contextual information—such as objects, people, body poses, and interactions—thereby enhancing gaze prediction and improving generalization to unseen scenarios \cite{gupta2024exploring}. This highlights the potential of VLMs to enhance contextual awareness in gaze modelling. For example, understanding that “person A is pointing at an object” or “person B is speaking” provides valuable context for predicting where person C might look \cite{gupta2024exploring}.

\subsection{GazeVLM}
Figure \ref{fig:gaze_vlm_intro} presents a high-level comparison between the existing methods and our approach. Building on the success of VLMs, we introduce GazeVLM, a novel approach to gaze understanding using VLM. Fine-tuned from Qwen2.5-VL\cite{Qwen2-VL}, GazeVLM generalizes across person detection, gaze target detection, and gaze object identification. Its architecture includes a vision tower to extract high-dimensional embeddings from RGB images and HHA-encoded (\textbf{H}orizontal disparity, \textbf{H}eight and \textbf{A}ngle) depth maps, followed by a cross-attention fusion module that integrates visual and geometric cues. The fused representation is processed by a text decoder to generate task-specific predictions, enabling robust gaze analytics. Unlike prior methods relying on visual features alone, GazeVLM’s multimodal design enhances accuracy and robustness, particularly in challenging scenarios like occlusions, out-of-frame gaze targets, and ambiguous contexts.

\subsection{Our contribution}

This paper introduces a novel approach for gaze understanding. Our key contributions are as follows:

\begin{enumerate}
\item \textbf{The first VLM-Based Framework for Unified Gaze Understanding:} To our knowledge, this is the first work to incorporate a vision–language model for unified gaze understanding. Unlike prior single-task methods that rely on traditional single-branch and multi-branch fusion methods, GazeVLM supports a multi-task learning setup encompassing person detection, gaze target detection, and gaze object identification.

\item \textbf{Efficient Depth Modality Integration via HHA Encoding:}  
We represent depth maps as HHA encodings to ensure better compatibility with the pretrained vision tower. Our contribution lies in designing a novel approach that enables both RGB and depth modalities to be processed through the same frozen vision encoder.

\item \textbf{GazeVLM Dataset:} We introduce a novel dataset that is prepared and curated from datasets like GazeFollow\cite{nips15_recasens} and VideoAttentionTarget\cite{chong2020detectingattendedvisualtargets} for  VLM adaptation. GazeVLM dataset is made available to the research community to facilitate further progress in this area.
\end{enumerate}

In the following sections, we outline the structure of this paper: Section \ref{sec:related} reviews prior work in gaze understanding and multi-modal VLM applications. Section \ref{sec:dataset} discusses dataset preparation for GazeVLM. Section \ref{methodology} elaborates on our multimodal approach and training methodology. Section \ref{experimentations} evaluates GazeVLM on \textit{gaze target detection} and \textit{gaze object identification} tasks. Finally, Section \ref{conclusion} provides concluding remarks and Section \ref{Limitations} identifies future directions.

\section{Related Work}
\label{sec:related}
\subsection{Single and multi-branch methods}

Gaze target detection has been predominantly addressed through multi-branch fusion models \cite{wang2024, Wang2022, Tafasca2024, tonini2023, Horanyi2023, Jin2023, liu2023}, where an encoder is followed by parallel branches designed to extract specific gaze-related cues. In contrast, single-branch methods \cite{ryan2024, jindal2024, Wang2023} utilize a feature extractor or visual encoder followed by a gaze regression module, prioritizing simplicity but potentially missing nuanced cues. Recent transformer-based architectures, such as TransGOP\cite{wang2024}, enhance gaze-object relationship modelling by combining an object detector, a gaze autoencoder, and an object-to-gaze cross-attention mechanism. While this approach demonstrates improved heatmap regression, its performance remains heavily reliant on the accuracy of the underlying object detector. Similarly, GaTector \cite{Wang2022} faced challenges in accurately localizing gaze boxes due to limitations in its object detection capabilities, highlighting the critical role of robust object detection in gaze estimation tasks. UVAGaze \cite{liu2023} improved performance through multi-view unsupervised gaze estimation but faced challenges with occlusion and deformation. On the other hand, Liang \textit{et al.} \cite{Liang2025} addressed data scarcity by combining multiple datasets across multiple branches, achieving
improved generalization. Despite its benefits, this approach faced limitations due to inconsistencies in annotations across datasets.

Recently, single-branch gaze estimation methods \cite{jindal2024, ryan2024, Haldun2023, tu2022end} have received limited attention in research. Jindal \textit{et al.} \cite{jindal2024} proposed an attention mechanism that dynamically adapts to spatial variations across sequential frames and incorporates a bias correction model using pre-trained Gaussian processes. However, this method struggles with long-term spatial and temporal dynamics. Similarly, Gaze-LLE \cite{ryan2024} employs a single-branch DINOv2 encoder followed by a head-prompting decoder. Although computationally efficient, its performance depends heavily on the encoder's size and quality, and it fails to capture diverse gaze-related cues such as head pose, eye movement, and contextual scene information, leading to lower accuracy. Another study \cite{tu2022end} introduced a transformer-based model to address the limitations of using head images as input. While achieving promising results, their approaches suffer from slow convergence when focusing on sparse, meaningful locations.

\subsection{Single and multi-modal methods}

Multi-modal methods \cite{miao2022, mondal2023, 2023childplay, Gupta2022, hou2023, Sood2023, mathew2023, Chen2022} typically follow a two-stage approach. In the first stage, task-specific predictors extract additional cues, such as depth and pose, to compensate for missing human-scene interactions. In the second stage, the extracted visual and multimodal features are combined, and a decoder predicts the gaze target heatmap. For instance, Athul \etal \cite{mathew2023} proposed a method that fuses 3D depth estimation, saliency mapping, and multimodal features. However, their approach performs frame-wise predictions and does not account for temporal dependencies. Chen \etal\cite{Chen2022} introduced a gaze-assisted object referring framework that simplifies existing methods by integrating gaze heatmaps with one-stage object detection and language support. Despite its innovation, their approach was computationally inefficient and struggled to handle diverse imaging modalities. Gupta \etal \cite{Gupta2022} addressed this limitation by extracting depth and pose maps from input images and combining them into modality saliency feature maps. While effective, their method incurs increased latency, impacting real-time performance.

Unlike multi-modal methods, single-modal approaches \cite{Shi2024, Wang2022, Haldun2023, Zhang2022GazeOnce} often suffer from limited feature representations. ViTGaze \cite{song2024vit} introduces a single-modality gaze-following framework that leverages vision transformers (ViTs) to model human-scene interactions through self-attention. Unlike traditional two-stage or complex decoder-based approaches, it focuses on powerful encoders with minimal decoder parameters, simplifying the architecture. To address challenges such as head pose variation and illumination, Shi \textit{et al.} \cite{Shi2024} proposed the Agent-guided Gaze Estimation Network (AGE-Net). This framework uses a main branch to capture eye features from low-level semantics, while agent regression tasks refine asymmetry using high-level semantics. Haldun \textit{et al.} \cite{Haldun2023} introduced U-Net Frame-to-Gaze, a method that predicts 3D gaze origin and direction directly from raw RGB camera frames without requiring eye or face cropping. However, it relies on prior knowledge of camera-to-screen geometry and suggests few-shot learning to adapt to new setups with minimal user input. ESCNet \cite{9878884} explicitly models 3D geometry by reconstructing the entire scene from a single RGB image and incorporates 3D gaze and scene contextual cues for predictions. While effective, this method requires high input resolution for accurate 3D scene reconstruction. 
 
\subsection{Vision language-based methods}
Vision Language Models (VLMs) have demonstrated impressive zero-shot performance across various vision tasks. However, accurate gaze following requires capturing contextual cues specific to each individual in the scene. Recent work \cite{gupta2024exploring} explores the use of VLMs for extracting zero-shot contextual cues to improve gaze following. Among these, BLIP-2 \cite{BLIP-2} achieves the best performance, leveraging in-context learning and prompt ensembling to enhance robustness. Using full images with an ellipse around the target improves visual prompting, leading to better generalization. However, the set of cues considered is fixed and depends on the chosen model. Gazeformer \cite{mondal2023} introduced ZeroGaze, a novel zero-shot gaze prediction task for unseen objects, and a transformer-based model that utilizes natural language encoding instead of object detectors. With five times faster performance, Gazeformer jointly embeds visual semantic features within language model features to predict human scanpaths for visual search. Indeed, VLMs have shown strong performance in many human-centric tasks, such as action recognition, person re-identification, and search \cite{wang2024Hulk}. However, only a few works have addressed gaze understanding. For instance, Yang \textit{et al.} \cite{jin2024} tackle gaze object prediction from an image segmentation perspective by leveraging pixel-level supervision provided vision foundation model. Their method relies on box supervision as prompts for SAM to generate masks, restricting its applicability in scenarios without strong location priors.

\begin{figure}[!ht]
  \centering
   \includegraphics[width=1.0\linewidth]{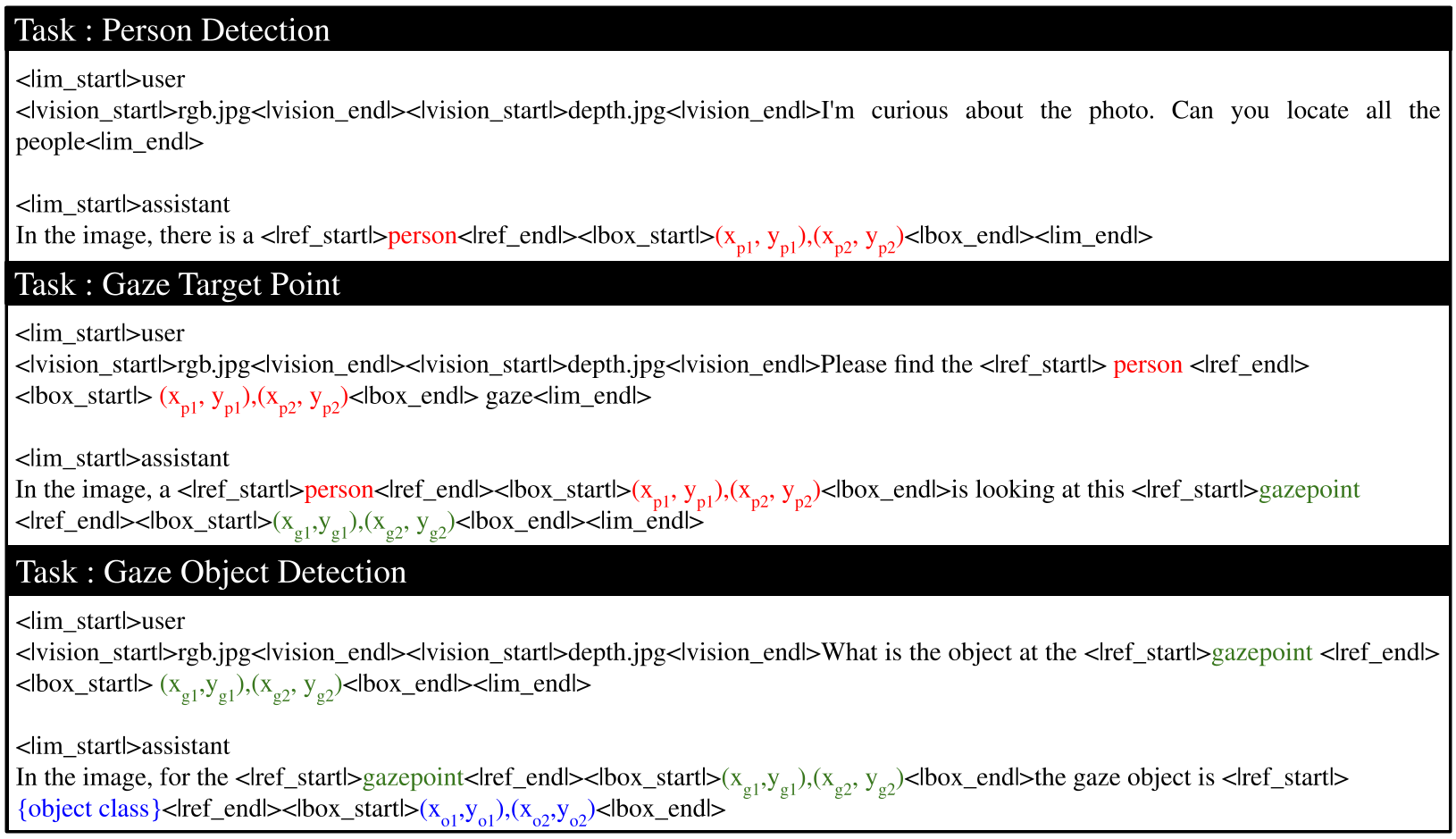}
   \caption{\textbf{Dataset format example.} Each task statement is marked with special tokens \texttt{<im\_start>} and \texttt{<im\_end>}. Image features are separated from text features using special tokens \texttt{<vision\_start>} and \texttt{<vision\_end>}, in line with ChatML\cite{openai_chatml} format.}
   \label{fig:datasetformat}
\end{figure}

\section{Dataset}
\label{sec:dataset}
GazeVLM addresses key tasks for third-person human-centric gaze systems, including person detection, gaze point regression, and gaze object localization and classification. It is trained and evaluated on two widely used datasets: GazeFollow\cite{nips15_recasens} and VideoAttentionTarget\cite{chong2020detectingattendedvisualtargets}.

GazeFollow is a large-scale static image dataset with 122,143 images and 130,339 head-target annotations, including a test set of 4,782 images. Each gaze target is annotated by ten individuals, ensuring robust ground truth. The dataset focuses primarily on in-frame gaze targets. VideoAttentionTarget extends gaze detection to video sequences, containing 1,331 clips 109,574 in-frame gaze targets and 54,967 out-of-frame gaze annotations. Its test set includes 31,978 gaze annotations from 10 different shows. The dataset is notable for its complexity, featuring multiple head-target annotations per frame, making it ideal for modelling multi-person gaze interactions.

Together, these datasets provide a comprehensive foundation for gaze target detection research across static and dynamic scenarios, significantly advancing the field and serving as valuable resources for researchers.

\subsection{Data Preparation}
\label{sec:data_preparation}

In our approach to leverage datasets for vision-language models, we transformed the original annotations into a text format suitable for the GazeVLM architecture. The GazeFollow dataset contains free-style images with multiple people, where the gaze of selected individuals is evaluated as part of the test set. To estimate the gaze of a specific person, we embed the face location of that individual into the text prompt. Additionally, to identify the gazed object, we extend the GazeFollow dataset with object annotations. We employ Detic\cite{zhou2022detectingtwentythousandclassesusing} object detection model pre-trained on the LVIS dataset\cite{gupta2019lvis}, which provides a rich vocabulary of 1200 object classes. This choice, over a closed-vocabulary detector, allows us to identify a wider and more diverse range of potential gaze targets without being limited to a predefined, fixed set of object categories. This is particularly beneficial in unconstrained real-world images where the gazed objects can be varied and novel. The gazed object is determined by computing the Intersection over Union (IoU) between the gaze point and each detected object:

\begin{equation}
\text{IoU} = \frac{\text{area}(B_{\text{gaze}} \cap B_{\text{obj}})}{\text{area}(B_{\text{gaze}} \cup B_{\text{obj}})}
\end{equation}

where $B_{\text{gaze}}$ is a small bounding box centered on the gaze point, and $B_{\text{obj}}$ is the bounding box of a detected object. The object with the highest IoU score is considered the gazed object:

\newcommand*{\argmax}{\mathop{\mathrm{arg\,max}}\limits}

\begin{equation}
O_{\text{gazed}} = \argmax_{o \in O_{\text{det}}} \text{IoU}(B_{\text{gaze}}, B_o)
\end{equation}

This approach associates gaze points with specific objects, thus enhancing the semantic understanding of gaze.

To improve the representation of bounding boxes and gaze points in VLMs, we propose an enhanced format based on the convention introduced by \cite{bai2023}. Our approach formalizes bounding boxes in the $xyxy$ configuration, normalized to the range $[0,1000)$, and represents them as strings:

\begin{equation}
B_{xyxy} = \texttt{``$(x_1, y_1), (x_2, y_2)$''}
\end{equation}

For gaze point estimation, we introduce a transformation to create a bounding box around the gaze point:

\begin{equation}
G_{(x,y)} \rightarrow B_{gaze} = \texttt{``$(x-\lambda, y-\lambda), (x+\lambda, y+\lambda)$''}
\end{equation}

where \( G(x,y) \) is a gaze point, and \( \lambda \) is a fixed margin. To enhance the model's understanding of spatial representations, we introduce special tokens for formatting bounding boxes:

\begin{equation}
B_{\text{token}} = \texttt{<box\_start>} B_{xyxy} \texttt{<box\_end>}
\end{equation}

This formatting scheme is applied uniformly to both object bounding boxes (\( B_{xyxy} \)) and gaze bounding boxes (\( B_{gaze} \)). For associating gazing objects with their corresponding classes, we utilize additional tokens:

\begin{equation}
O_{token} = \texttt{<ref\_start>} O_{class} \texttt{<ref\_end>} B_{token}
\end{equation}

where \( O_{class} \) is the object class label. The complete representation for a gaze estimation scenario is:

\begin{equation}
S_{gaze} = \texttt{<box\_start>} B_{gaze} \texttt{<box\_end>} + O_{token}
\end{equation}

Given Qwen2.5-VL's inherent inability to directly predict spatial coordinates for grounding, we employ the above formulations to comply with its training data format, thereby facilitating the fine-tuning process for gaze understanding tasks. An example for the dataset format is shown in Figure \ref{fig:datasetformat}. Additionally, in order to complement GazeVLM with depth understanding, we use monocular depth estimation network from \cite{Ranftl2022} to extract the depth maps from the RGB images.

\section{The Methodology}
\label{methodology}
Our work focuses on the task of gaze-related understanding in images, which encompasses three key components: person detection, gaze target detection, and gaze object identification. Given an RGB image \( I_{\text{RGB}} \in \mathbb{R}^{H \times W \times 3} \) and its corresponding depth map \( D \in \mathbb{R}^{H \times W} \), our goal is to predict the gaze target location \( (x, y) \) within the image and identify the object being gazed at, even when the object lies outside the image. This formulation allows our model to handle a wide range of real-world scenarios.

\subsection{Input Representation}
The input to our model consists of two modalities: an rgb image and a depth image. The rgb image provides rich visual information about the scene, while the depth map, encodes geometric information about the relative distances of objects in the scene. Drawing inspiration from \cite{gupta2014learningrichfeaturesrgbd},we feed HHA geocentric images rather than directly inputting raw depth maps into the vision encoder of the VLM. HHA encoding comprises three components: \textbf{Horizontal Disparity (H)}, \textbf{Height (H)}, and \textbf{Angle (A)}, representing inverse depth, vertical pixel position, and surface orientation relative to the camera, respectively. As discussed in Section \ref{sec:ablation_study}, our rationale is that since vision encoders are predominantly pretrained on RGB images, converting the depth map into a three-channel HHA encoding enhances its compatibility with these pretrained encoders. The structural similarities between HHA geocentric images and RGB images allow the vision encoder, originally designed for RGB data, to effectively extract complementary representations from the HHA images. Our solution enables the extraction of depth-related representations using the vision encoder without the necessity of additional pretraining. Below, we outline the steps to compute the HHA encoding from a depth map \( D \in \mathbb{R}^{H \times W} \).

\subsubsection{Depth Map Rescaling}
First, the depth map \( D \) is rescaled to the range \([1, 10]\) to avoid extreme values in the inverse depth computation:
\begin{equation}
D_{\text{rescaled}} = \text{normalize}(D, 1, 10)  
\end{equation}
where \(\text{normalize}(\cdot)\) linearly maps the input to the the specified range.

\begin{figure}[!ht]
  \centering
   \includegraphics[width=1.0\linewidth]{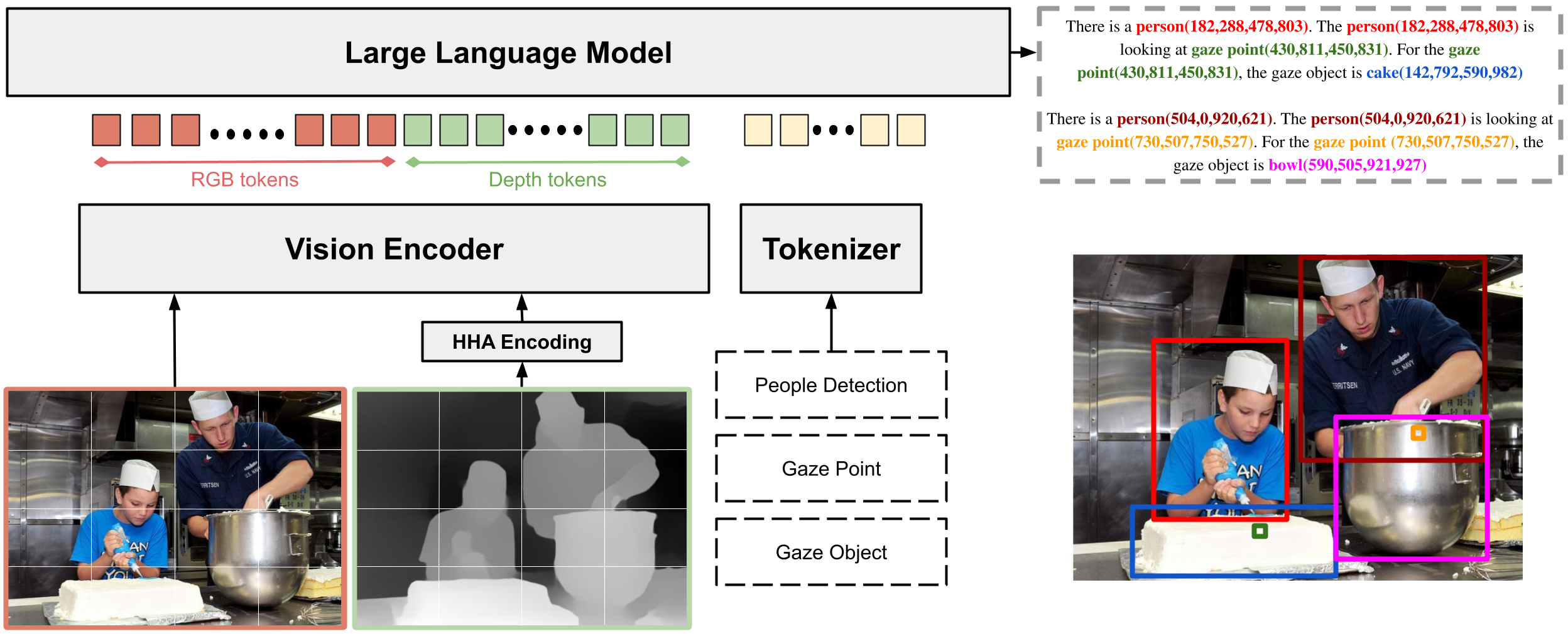}
   \caption{\textbf{Overview of GazeVLM.} The model processes an input comprised of an RGB image and a corresponding HHA-encoded depth map. Our model is multi-task and can detects individuals, their gaze points, and the objects they focus on, based on the task provided by the user input prompt.}
   \label{fig:gazevlm}
\end{figure}

\subsubsection{Horizontal Disparity (H)}
The horizontal disparity channel \( H \) is computed as the inverse of the rescaled depth map. To ensure numerical stability, zero values in \( D_{\text{rescaled}} \) are replaced with a small constant \( \epsilon = 10^{-6} \). The resulting \( H \) is normalized to the range \([0, 255]\):
\begin{equation}
H = \text{normalize}(1 / D_{\text{rescaled}}, 0, 255)
\end{equation}

\subsubsection{Height (H)}
The height channel \( H_{\text{height}} \) is computed based on the vertical position of each pixel in the image:
\begin{equation}
H_{\text{height}} = \left( \frac{y}{H} \right) \times 255
\end{equation}
where \( y \) is the vertical coordinate of the pixel, and \( H \) is the height of the image.

\subsubsection{Surface Normals and Angle (A)}
The surface normals \( \mathbf{n} \) are computed from the depth map using Sobel operators for gradient computation:
\begin{equation}
\frac{\partial D}{\partial x} = \text{Sobel}(D, \text{axis}=0), \quad
\frac{\partial D}{\partial y} = \text{Sobel}(D, \text{axis}=1)
\end{equation}
\begin{equation}
\mathbf{n} = \left[ -\frac{\partial D}{\partial x}, -\frac{\partial D}{\partial y}, 1 \right]
\end{equation}
The normals are then normalized to unit length:
\begin{equation}
\mathbf{n}_{\text{normalized}} = \frac{\mathbf{n}}{\|\mathbf{n}\|_2}
\end{equation}

The angle channel \( A \) is computed as the angle between the surface normal \( \mathbf{n}_{\text{normalized}} \) and the camera's viewing direction \( \mathbf{v} = [0, 0, 1] \). The resulting \( A \) is normalized to the range \([0, 255]\):
\begin{equation}
A = \text{normalize}(\arccos(\text{clip}(\mathbf{n} \cdot \mathbf{v}, -1, 1)), 0, 255)
\end{equation}

\subsubsection{HHA Encoding}
Finally, the simplified HHA encoding is obtained by stacking the three channels:
\begin{equation}
\text{HHA} = \text{stack}(H, H_{\text{height}}, A)
\end{equation}

\subsection{Model Architecture}
GazeVLM is built on top of Qwen2.5-VL, which we fine-tune for gaze-related tasks. The model architecture can be seen in Figure \ref{fig:gazevlm}. It consists of two main components: a vision tower and a text decoder. The vision tower processes the RGB image and the HHA-encoded depth map independently, extracting high-dimensional feature embeddings for each modality. These embeddings alongside text embeddings are then fused using a cross-attention mechanism, which allows to effectively combine visual, geometric and textual information. 

\subsection{Fusion Strategy}

To integrate multimodal inputs, the RGB image ($F_{\text{RGB}}$) and HHA-encoded depth map ($F_{\text{HHA}}$) are independently processed by the vision tower. HHA encoding is adopted to align depth information with the RGB-centric Qwen2.5-VL encoder. These visual embeddings are fused with text embeddings ($F_{\text{text}}$) via cross-attention in the LLM decoder:$$F_{\text{fused}} = \text{Softmax}\left(\frac{QK^T}{\sqrt{d_k}}\right)V$$where queries ($Q$) are projected from $F_{\text{text}}$, and keys ($K$) and values ($V$) are derived from the concatenated visual features $F_V = [F_{\text{RGB}}; F_{\text{HHA}}]$. By computing attention between the textual query and the joint visual representation, the model dynamically weights appearance cues from RGB and geometric structures from HHA. This allows the decoder to extract complementary information from either modality based on the specific requirements of the text prompt.

\subsection{Training Procedure}
We fine-tune the model with GazeFollow and VideoAttentionTarget datasets. Before training, we preprocess the data by converting depth maps to HHA encoding and formatting the datasets to match the input requirements of Qwen2.5-VL as outlined in Section \ref{sec:data_preparation}. We use the 3B variant of Qwen2.5-VL (Qwen2.5-VL-3B-Instruct). AdamW optimizer with a learning rate of \( 1e-5 \) and gradient clipping is applied to ensure stable optimization. Given image $I$ and user input text prompt $Q$, the model generates answer $A = \{a_1, ..., a_n\}$. The fine-tuning minimizes the Negative Log-Likelihood of the answer tokens:
\begin{equation}
\mathcal{L} = - \sum_{i=1}^{n} \log P(a_i | I, Q, a_{<i})  
\end{equation}

where $P(a_i | I, Q, a_{<i})$ is the probability of token $a_i$ given the input and preceding tokens.

\begin{figure}[ht]
  \centering
   \includegraphics[width=1.0\linewidth]{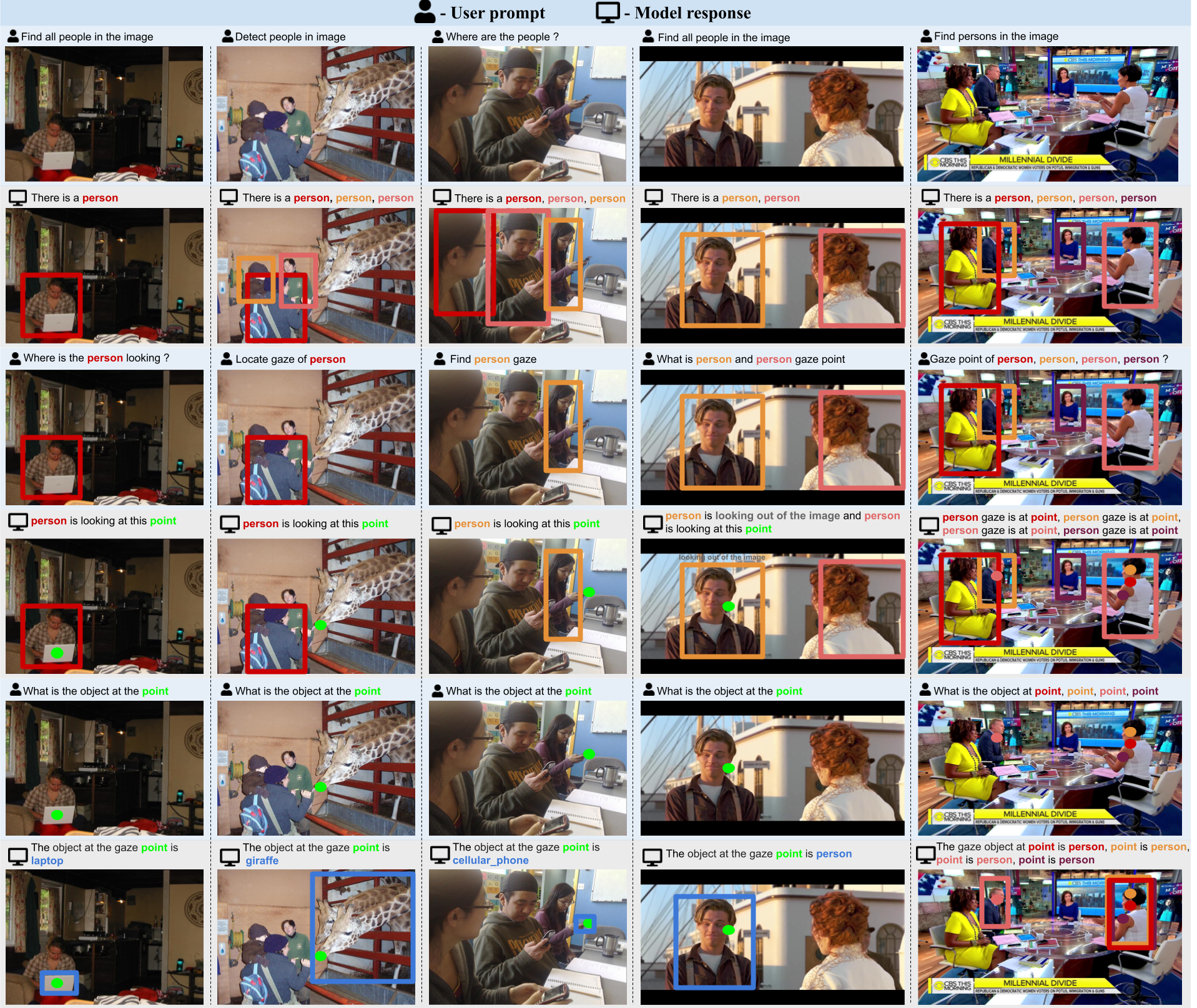}
   \caption{\textbf{Qualitative results from GazeFollow and VideoAttentionTarget datasets.} Each column denotes an example image in a series of multi-turn user input prompts and model responses. User input prompt and model response for every image is highlighted row-wise using \faUser\ and \faDesktop\ icons respectively. The example in fourth column also demonstrates a scenario for gaze in/out classification. The model responds with a textual tag \textit{"looking out of the image"} if the person gaze is not within the field of view of the scene image. All model responses are spatially located and color-coded for easier visualization.}
   \label{fig:gazevlm_examples}
\end{figure}

\section{Experimentation}
\label{experimentations}

\subsection{Evaluation Metrics}
The evaluation of GazeVLM was conducted across two fundamental gaze-related tasks: \textbf{gaze target detection} and \textbf{gaze object identification}. The Area Under the Curve (AUC) metric assesses the confidence of the predicted gaze heatmap with respect to the ground-truth gaze data, quantifying how well the predicted heatmap aligns with the actual gaze distribution. The Distance (Dist.) metric calculates the L2 distance (Euclidean distance) between the ground-truth gaze point and the predicted gaze location. To evaluate directional accuracy, the Angle metric is used to measure the angular deviation between the predicted gaze vector and the ground-truth gaze vector. Specifically for the GazeFollow dataset, the minimum Euclidean distance between the predicted gaze point and the ten annotated ground-truth target points for each subject was calculated. In datasets such as VideoAttentionTarget, which include out-of-frame gaze targets, the model's ability to distinguish between in-frame and out-of-frame gaze was evaluated by calculating the average precision (AP) for out-of-frame gaze probability predictions.

For gaze object detection, the Mean Average Precision (mAP) metric was adopted to assess the model's performance in object class detection and localization. A prediction is considered correct if the predicted bounding box’s class label matches the ground-truth class label and the Intersection over Union (IoU) between the predicted bounding box and the ground-truth bounding box exceeds a predefined threshold. The IoU, quantifying the spatial overlap between the predicted and ground-truth bounding boxes, was predominantly set to 0.5, reflecting a standard evaluation protocol.

\subsection{GazeVLM Evaluation}
A comparative analysis against state-of-the-art architectures is shown in Table \ref{tab:results_bmvc}. The method \textit{Random} samples gaze predictions from a standard normal distribution, while \textit{Center} assumes predictions at the image center. \textit{Fixed Bias} reflects dataset biases in face position and gaze fixation points. GazeVLM achieves state-of-the-art scores accross all metrics on GazeFollow and VideoAttentionTarget datasets. A significant contribution of this work is the ability of GazeVLM to perform gaze object detection, achieving a competitive mean Average Precision (mAP) metric score of 0.25 on VideoAttentionTarget and 0.23 on GazeFollow datasets.
 

Figure \ref{fig:gazevlm_examples} illustrates GazeVLM's ability to execute diverse tasks based on specific user prompts. Unlike traditional gaze estimation frameworks that rely on auxiliary networks for head detection, GazeVLM employs a unified, self-contained architecture that achieves performance comparable to, or exceeding, specialized state-of-the-art models. By eliminating the need for external components and auxiliary networks, GazeVLM provides a streamlined, stand-alone solution for multiple gaze-related downstream tasks.

\begin{table*}[!ht]
\begin{center}
\setlength{\tabcolsep}{8pt}
\begin{tabular}{|l|cccc|ccccc|}
\hline
 & \multicolumn{4}{c|}{\textbf{VideoAttentionTarget}} & \multicolumn{5}{c|}{\textbf{GazeFollow}} \\
\cline{2-5} \cline{6-10}
Method & $AUC$ $\uparrow$ & $Dist.$ $\downarrow$ & $AP$ $\uparrow$& $AP_{ob}$ $\uparrow$ & $AUC$ $\uparrow$ & $Dist.$ $\downarrow$ & $M. Dist.$ $\downarrow$ & $Angle$ $\downarrow$ & $AP_{ob}$ $\uparrow$ \\
\hline
Random & 0.505 & 0.458 & 0.621 & - & 0.504 & 0.484 & 0.391 & 69.0 & - \\
Center & - & - & - & - & 0.633 & 0.313 & 0.230 & 49.0 & - \\
Fixed Bias & 0.728 & 0.326 & 0.624 & - & 0.674 & 0.306 & 0.219 & 48.0 & - \\
Recansens \cite{nips15_recasens} & - & - & - & - & 0.878 & 0.190 & 0.113 & 24.0 & - \\
Chong \cite{Chong2018} & 0.830 & 0.193 & 0.705 & - & 0.896 & 0.187 & 0.112 & - & - \\
Lian \cite{Lian2019} & 0.837 & 0.165 & - & - & 0.906 & 0.145 & 0.081 & 17.6 & - \\
Danyang \cite{tu2022end} & 0.893 & 0.137 & 0.821 & - & 0.917 & 0.133 & 0.069 & - & - \\
VideoAttn \cite{chong2020detectingattendedvisualtargets} & 0.860 & 0.134 & 0.853 & - & 0.921 & 0.137 & 0.077 & - & - \\
Fang \cite{9577574} & 0.905 & \underline{0.108} & 0.896 & - & 0.922 & 0.124 & 0.067 & 14.9 & - \\
Jin \cite{tu2022end} & 0.870 & 0.127 & 0.882 & - & 0.923 & 0.120 & 0.064 & 14.8 & - \\
Tonini \cite{Tonini_2022} & \underline{0.940} & 0.129 & - & - & 0.927 & 0.141 & - & - & - \\
Bao \cite{9878884} & 0.885 & 0.120 & 0.869 & - & 0.928 & \textbf{0.122} & - & \underline{14.6} & - \\
Miao \cite{miao2022} & 0.917 & 0.109 & \textbf{0.908} & - & 0.934 & \underline{0.123} & 0.065 & - & - \\
Tafasca \cite{2023childplay} & 0.914 & 0.109 & 0.834 & - & \textbf{0.939} & \textbf{0.122} & \underline{0.062} & - & - \\
\textbf{GazeVLM} & \textbf{0.948} & \textbf{0.102} & \underline{0.898} & \textbf{0.25} & \underline{0.936} & \underline{0.123} & \textbf{0.061} & \textbf{14.2} & \textbf{0.23} \\
\hline
\end{tabular}
\end{center}
\caption{Evaluation of VideoAttentionTarget and GazeFollow datasets. $AP_{ob}$ is computed for 1200 LVIS vocabulary classes at threshold=$0.5$. The best and second-best for each metric is represented in \textbf{bold} and \underline{underline}.}
\label{tab:results_bmvc}
\end{table*}

\subsection{Ablation Study}
\label{sec:ablation_study}

To evaluate the impact of HHA depth encoding, we compared three GazeVLM configurations: RGB-only, RGB + depth, and RGB + HHA. As shown in Table \ref{tab:results2}, the performance of GazeVLM varies across these configurations on the GazeFollow dataset. Interestingly, incorporating raw depth maps with GazeVLM's frozen vision encoder significantly degraded performance compared to the RGB-only baseline. The inclusion of HHA geocentric images provides essential supplementary information \cite{gupta2014learningrichfeaturesrgbd} that enhances GazeVLM's performance. The observation of performance degradation when raw depth maps are used with GazeVLM's frozen vision encoder is unlike conventional multi-stage networks where depth encoders consistently improved overall accuracy. Pretraining the vision encoder with depth maps could allow the direct use of raw depth maps for better accuracy. However, these results demonstrate that while direct depth integration may require expensive pre-training, HHA provides a computationally efficient alternative that captures essential geometric cues to complement RGB features.

\begin{table}[ht]
\begin{center}
  \begin{tabular}{|l|ccc|}
   \hline
    Method & AUC $\uparrow$ & Dist. $\downarrow$ & Angle $\downarrow$ \\
    \hline\hline
    \textbf{GazeVLM} (RGB only) & 0.928 & 
 0.069 & 14.5 \\
     \textbf{GazeVLM} (RGB+depth) & 0.921 & 
 0.073 & 15.1 \\
    \textbf{GazeVLM} (RGB+HHA) & 0.936 & 0.061 & 14.2 \\
    \hline
  \end{tabular}
  \end{center}
  \caption{Comparison of GazeVLM performance under varying configurations of input modalities.}
  \label{tab:results2}
\end{table}

\section{Conclusion}
\label{conclusion}
GazeVLM represents a significant step towards more accurate and robust gaze understanding. By leveraging the power of VLMs, GazeVLM can effectively integrate vision and text modalities to infer the target of a person's gaze. This has the potential to revolutionize human-computer interaction, enabling more natural and intuitive communication between humans and machines. The innovative approach and methodology employed in this research position GazeVLM as a promising solution for diverse applications across AI, robotics, and behavioural research.

\section{Limitations and Further Work}
\label{Limitations}
Currently, GazeVLM supports only static image-based analysis. It lacks video-based gaze understanding, which limits its ability to capture temporal dynamics. Further work will extend GazeVLM to videos by incorporating visual prompts to focus on regions of interest across frames. In addition, our goal is to improve the computational efficiency of GazeVLM for real-time applications. We plan to explore optimization mechanisms to enhance its practicality for real-world use.

\bibliography{references}

\end{document}